# Multimodal Large Language Models for Bioimage Analysis

Shanghang Zhang[1]✉, Gaole Dai[1], Tiejun Huang[1] & Jianxu Chen[2]✉

[1]National Key Laboratory for Multimedia Information Processing, School of Computer Science, Peking University, Beijing, China
[2]Leibniz-Institut für Analytische Wissenschaften – ISAS – e.V., Dortmund, Germany
✉shanghang@pku.edu.cn, jianxu.chen@isas.de

Rapid advancements in imaging techniques and analytical methods over the past decade have revolutionized our ability to comprehensively probe the biological world at multiple scales, pinpointing the type, quantity, location, and even temporal dynamics of biomolecules. The surge in data complexity and volume presents significant challenges in translating this wealth of information into knowledge. The recently emerged Multimodal Large Language Models (MLLMs) exhibit strong emergent capacities, such as understanding, analyzing, reasoning, and generalization. With these capabilities, MLLMs hold promise to extract intricate information from biological images and data obtained through various modalities, thereby expediting our biological understanding and aiding in the development of novel computational frameworks. Previously, such capabilities were mostly attributed to humans for interpreting and summarizing meaningful conclusions from comprehensive observations and analysis of biological images. However, the current development of MLLMs shows increasing promise in serving as intelligent assistants or agents for augmenting human researchers in biology research (Fig.1a).

## Multimodal large language models

The "scaling law" of foundation models, which refers to the observed trend that the performance of downstream tasks generally improves as the size of models increases (both in terms of training data and parameters), implies that the models can achieve generalization capability and even emergent ability when they exceed a certain scale[1]. Large language models (LLMs) are a type of foundation model that were originally developed to comprehend, generate, and manipulate diverse linguistic tasks through a comprehensive and expansive model. Benefiting from the "scaling law", LLMs achieve strong generalization capabilities across various downstream tasks. However, since the real-world data exist in different modalities, not only in language, there has been a growing trend toward utilizing LLMs as the basis for multimodal learning[2], leading to multimodal large language models (MLLMs). MLLMs effectively encode inputs from various modalities such as text, image, and audio simultaneously, enabling the transformation of these inputs into uniform features, followed by a fusion process to facilitate information integration across modalities, and at the end employ LLMs such as Generative Pre-trained Transformers (GPTs)[3] to decode the fused features into comprehensive outputs, such as analyzing, summarizing, reasoning, or even planning.

The great complexity and quantity of bioimage data bring common problems in current bioimage analysis. Bioimage data are usually faced with various conditions, domains, tasks, and even modalities, while it is prohibitively expensive to annotate enough data for all the different scenarios. How to design a method that can automatically generalize to various scenarios is an important and challenging problem in bioimage analysis. Besides, it usually requires a lot of human expertise and effort to monitor and analyze the bioimaging process, exceeding the capability of individual expert. How to build a model with wide spectrum of knowledge to automatically analyze bioimages is also a challenging problem. MLLMs with their generalization capabilities and emergent abilities enable a potentially next-generation paradigm for bioimage analysis. MLLMs demonstrate superior generalization capabilities across various downstream tasks compared to pure vision models, such as Segment Anything (SAM)[4], which only have strong abilities in the specific learned task. For example, an MLLM trained to identify nuclei in wild-type, can still accurately recognize nuclei even after being exposed to different compounds by providing the specific treatment information through a guiding text prompt. In contrast, pure vision models fail to exhibit this capability. Furthermore, larger models can exhibit emergent capabilities - the ability to perform tasks well that they were not explicitly programmed to do. This arises from the models' increased capacity to learn complex patterns and relationships within the training data, a feature absents in smaller models. This means an MLLM with emergent abilities can integrate wider spectrum of knowledge and diverse information to offer an intelligent and automatic way to analyze bioimages.

It is worth noting that in the context of biological studies, the term "multimodal" may have slightly different definitions. In biology, multimodal integration typically refers to combining data from two or more complementary modalities, e.g., different imaging techniques[5] while in general machine learning this commonly refers to processing information from different data types beyond just texts, e.g., images, speech or videos.

## Existing attempts for building MLLMs for bioimage analysis

There have been few works directly building MLLMS for bioimage analysis, since it is still an emerging topic. Most related works leverage LLMs to enhance software interaction with humans. One milestone was the development of Omega[1], which leverages LLMs to enable users to define their tasks using plain English, and generate and execute necessary code for bioimage processing and analysis on the napari[2] platform. We believe such interaction is not restricted to bioimage analysis, but could also be extended to interactive acquisition and analysis, as conceptualized for the future of Smart Microscopes[6,7] (Fig.1b).

The development of core foundational models for various modalities in biology paves the path to the utilization of MLLMs for bioimage analysis. This includes not only foundation models for microscopy images, such as Universal Fluorescence Microscopy-based Image Restoration[8] or Segment Anything for Microscopy[9], but also for closely related modalities, such as scGPT[10] and nicheformer[11] for integrating single-cell omics data. The utilization of these powerful tools firmly supports that MLLMs should assume an indispensable role in facilitating comprehensive analysis by bridging biological images and omics data, showcasing their remarkable generalization capabilities across diverse modalities.(Fig.1b)

---

[1] [GitHub - royerlab/napari-chatgpt: A napari plugin to process and analyse images with chatGPT!](#)
[2] [GitHub - napari/napari: napari: a fast, interactive, multi-dimensional image viewer for python](#)

# Steps towards constructing MLLM systems for bioimage analysis

We aim to build intelligent MLLM systems for bioimage analysis that can not only accurately analyze the images, but also augment human researchers throughout the entire research process from designing imaging assays and acquiring data to knowledge discovery. MLLMS can be used for bioimage analysis from three aspects: (1) Directly using MLLMs for more accurate and robust bioimage analysis, such as performing image segmentation or anomaly detection (e.g., automatically finding low-quality images in high throughout experiments). MLLMs can provide a generalized image analyzing method that is adaptive to different domains and downstream tasks; (2) Using MLLMs for automatic report generation for large-scale bioimage analysis. Studies such as drug screening and cell profiling usually require a large amount of image data, making it difficult to do a traditional summary of the data. MLLMs can be used to understand both metadata (as language input) and large amount of bioimages to automatically generate analysis, summarizations, and reports; (3) Using MLLMs as agent for Smart Microscopes. By analyzing the microscopy image samples (e.g., blur or out-of-focus), MLLMs can generate action codes to adjust the microscope automatically to obtain better observations. To realize the above three aspects, we envision three major steps with extensive community efforts, described as "bricks, buildings, and facilities"(Fig1.c).

First, we need "bricks" to train MLLMs, i.e., constructing a substantial dataset comprising paired images and accompanying text descriptions, and even other modalities. The example of GPT-3[3], trained on a massive 45TB CommonCrawl dataset[12] (condensed to 600GB by a 100+ person labeling team), highlights the challenge of creating a biological multimodal dataset. Fortunately, the community effort in developing FAIR public data repository creates great potentials; for example, repositories like the Image Data Resource[3] (IDR) provides image datasets with well-structured metadata for public access, which are usually associated with peer-reviewed publications. Initial attempts like Visual and Linguistic Integration of BioRxiv (VLIB)[13] utilize paired figures and captions from pre-prints to construct MLLMs for cancer biology. Another promising direction is the utilization of electronic laboratory notebooks (ELNs). Connecting ELNs to FAIR data management tools can create large datasets with texts information paired with microscopy images, data from other modalities, and even analysis results.

The second stage focuses on the architectural design of the three key components in MLLMs: encoders, fusion and alignment modules, and LLM decoders. Thanks to the open-source effort in the AI communities, there are extensive "blueprints" that can be used for these "buildings". For example, the fusion and alignment module commonly employ a popular technique known as Mixture of Experts (MoE), demonstrably successfully in both close-source MLLMs (e.g., GPT-4V) and open-source models (e.g., the Mixtral family). This approach enables scalable model expansion while effectively leveraging multimodal features, similar to how distinct brain regions handle diverse information types in humans.

The final stage involves transforming the "buildings" from stage 2 into practically functional "facilities" by fine-tuning them to solve bioimage analysis problems and avoid hallucinations. We use MLLMs to act as an agent that invoke appropriate tools, e.g., sending code to micro-manager to adjust microscope settings or generating a summary report after automatically analyzing a large-scale image-based screening experiments, akin to humans for solving specific tasks. Challenges remain: how to ensure MLLM trustworthiness and their functionality on new concepts or unseen modalities. Two transformative techniques provide promising solutions: Retrieval Augmented Generation (RAG)[14] and Parameter

---

[3] IDR: Image Data Resource (openmicroscopy.org)

Efficient Fine-tuning (PEFT)[15]. RAG redefines the capability of MLLMs and works as guardrails. It ensures the validity of model outputs through a retrieval mechanism that dynamically searches existing information sources. For example, imagine an MLLM agent designed for two tasks, one producing automatic segmentation of mitochondria from confocal microscopy images of stem cells and calculating morphological features, while the other one evaluating the results. With RAG, the agent could search relevant literature and find the normal width of mitochondria in the same conditions. If the measured width falls outside the normal range, the agent can take the feedbacks to refine itself to generate better segmentation. PEFT plays a critical role when further training is required for a new domain or task to improve the performance and avoid hallucinations. PEFT allows efficient knowledge transfer by training only a small fraction of model parameters, making fine-tuning efficient and preventing overfitting with limited training data. This last stage is where MLLMs could finally make solid contribution to accelerate scientific discoveries and push beyond existing boundaries of knowledge.

Cutting-edge imaging and analytical methods generate multimodal data of great complexity and quantity, creating significant challenges in bioimage analysis. Effective solutions usually require a wide spectrum of knowledge: the biological context, microscope properties, image processing algorithms, machine learning techniques, statistics, software engineering skills, etc., surpassing the capability of individual or even a small team of experts. MLLMs with their ability to integrate diverse information, exceptional generalization skills, and unexpected emergent abilities offer a promising new avenue to overcome these hurdles. Here, we provide an example of how to use MLLM as agent for Smart Microscopes to solve "big" biological question: (1) start with designing a concrete actionable experimental assay, (2) invoke the robotic microscope(s) or analytical devices to execute, (3) iteratively adjust the imaging protocol according to the evaluation of trial acquisitions so that the acquired images and data will be optimized for downstream analysis (4) invoke file management systems to register the data and meta data, (5) iteratively optimize image analysis results, taking all relevant information into account, e.g., the optical properties of microscopes, existing knowledge of the biological samples, the downstream analysis goals, etc., (6) conduct necessary analysis and invoke specialized modeling or visualization software when necessary, (7) generate a summary report according to the big biological question. In the future, we believe bioimage analysis will not be tackled as a pure image processing problem any more, but via an intelligent bioimaging research MLLM agent (Fig1.b).

## Acknowledgements

S. Z. is supported by the National Science and Technology Major Project of China (No. 2022ZD0117801). J. C. is funded by the Federal Ministry of Education and Research (Bundesministerium für Bildung und Forschung, BMBF) in Germany under the funding reference 161L0272, and also supported by the Ministry of Culture and Science of the State of North Rhine-Westphalia (Ministerium für Kultur und Wissenschaft des Landes Nordrhein-Westfalen, MKW NRW).

## Competing interests

The authors declare no competing interests.

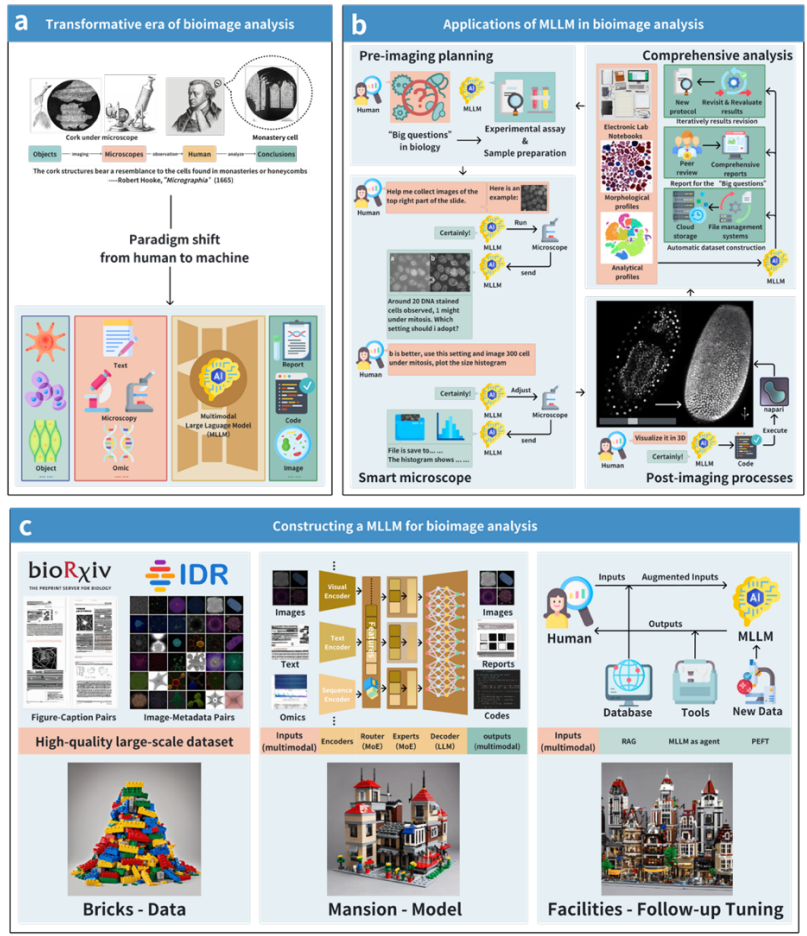

**Figure 1. MLLMs for biological image analysis. a.** Human experts have traditionally been responsible for gathering and interpreting meaningful conclusions from observations in bioimage analysis (top). However, the introduction of MLLM as a human assistant can simplify this complex task and enable the automation of research processes (bottom). **b.** Prior to imaging, MLLM can assist in initiating experimental assays, sample ordering and preparation. During imaging, MLLM acts as a surrogate for researchers by continuously adjusting microscope setups according to their needs, freeing them from long-term monitoring. After imaging, MLLM assists in visualizing and post-processing images by generating compatible codes for visualization platforms. Lastly, MLLM efficiently browses through vast datasets and summarizes potential answers to the initial big questions. **c.** Data from various modalities can be collected through platforms like BioRxiv and IDR, providing sufficient resources for constructing MLLMs (left). The architecture of an MLLM consists of multiple separate encoders with MoE blocks used for fusion; this allows decoding of fused features by LLM (middle). To further enhance the capabilities of MLLMs efficiently without requiring additional training, adjustments can be made using methods like RAG or as agents; when training is necessary, PEFT offers an efficient solution for incorporating new knowledge into MLLMs (right).